\documentclass[sigconf]{acmart}
\usepackage{graphicx}
\usepackage[export]{adjustbox}
\usepackage{cuted}
\usepackage{capt-of}

\AtBeginDocument{%
  \providecommand\BibTeX{{%
    \normalfont B\kern-0.5em{\scshape i\kern-0.25em b}\kern-0.8em\TeX}}}

\setcopyright{acmcopyright}
\copyrightyear{2023}
\acmYear{2023}
\acmDOI{XXXXXXX.XXXXXXX}

\acmConference[PCSC'23]{Philippine Computing Society Conference}{March 23--24,
  2023}{Cebu City, PH}
\acmPrice{15.00}
\acmISBN{978-1-4503-XXXX-X/18/06}

\begin{document}

\title{Baybayin Character Instance Detection}

\author{Adriel Isaiah V. Amoguis}
\email{adriel_amoguis@dlsu.edu.ph}
\affiliation{%
  \institution{De La Salle University}
  \city{Manila}
  \country{Philippines}}

\author{Gian Joseph B. Madrid}
\email{gian_joseph_madrid@dlsu.edu.ph}
\affiliation{%
  \institution{De La Salle University}
  \city{Manila}
  \country{Philippines}}
  
\author{Benito Miguel D. Flores, IV}
\email{benito_flores@dlsu.edu.ph}
\affiliation{%
  \institution{De La Salle University}
  \city{Manila}
  \country{Philippines}}

\author{Macario O. Cordel, II}
\email{macario.cordel@dlsu.edu.ph}
\affiliation{%
  \institution{De La Salle University}
  \city{Manila}
  \country{Philippines}}

\renewcommand{\shortauthors}{Amoguis, Madrid, Flores, \& Cordel}

\begin{abstract}
  The Philippine Government recently passed the "National Writing System Act," which promotes using \emph{Baybayin} in Philippine texts. In support of this effort to promote the use of \emph{Baybayin}, we present a computer vision system which can aid individuals who cannot easily read \emph{Baybayin} script. In this paper, we survey the existing methods of identifying \emph{Baybayin} scripts using computer vision and machine learning techniques and discuss their capabilities and limitations. Further, we propose a \emph{Baybayin} Optical Character Instance Segmentation and Classification model using state-of-the-art Convolutional Neural Networks (CNNs) that detect \emph{Baybayin} character instances in an image then outputs the Latin alphabet counterparts of each character instance in the image. Most existing systems are limited to character-level image classification and often misclassify or not natively support characters with diacritics. In addition, these existing models often have specific input requirements that limit it to classifying \emph{Baybayin} text in a controlled setting, such as limitations in clarity and contrast, among others. To our knowledge, our proposed method is the first \emph{end-to-end} character instance detection model for \emph{Baybayin}, achieving a mAP50 score of 93.30\%, mAP50-95 score of 80.50\%, and F1-Score of 84.84\%.
  
  
\end{abstract}

\keywords{optical character recognition, instance segmentation, computer vision, character classification, Baybayin}

\maketitle

\begin{center}
    \begin{figure}[t]
        \begin{tabular}{ccc}
            \textbf{Latin Label} & \textbf{Input Image} & \textbf{Output Image} \\
            
            GWAPO & 
            \includegraphics[width=.3\columnwidth]{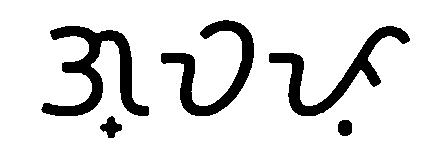} & 
            \includegraphics[width=.3\columnwidth]{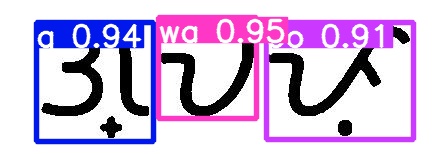} \\

            INIWAN &
            \includegraphics[width=.3\columnwidth]{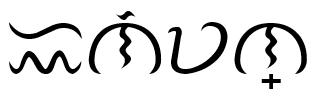} & 
            \includegraphics[width=.3\columnwidth]{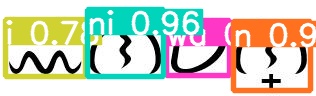} \\

            PILIPINO &
            \includegraphics[width=.3\columnwidth]{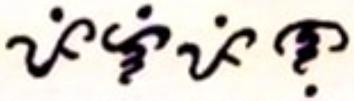} & 
            \includegraphics[width=.3\columnwidth]{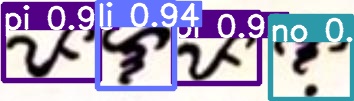} \\

            AKALA &
            \includegraphics[width=.3\columnwidth]{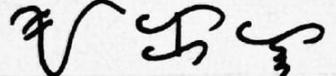} & 
            \includegraphics[width=.3\columnwidth]{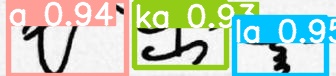} \\

            KAKALMA &
            \includegraphics[width=.3\columnwidth]{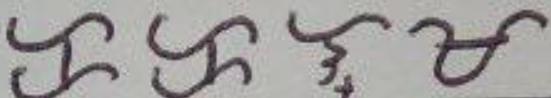} & 
            \includegraphics[width=.3\columnwidth]{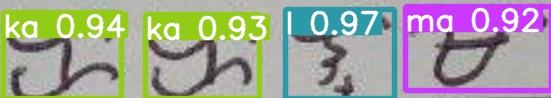} \\

            MAKUKUHA &
            \includegraphics[width=.3\columnwidth]{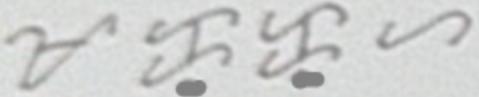} & 
            \includegraphics[width=.3\columnwidth]{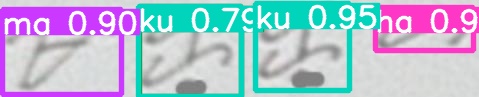} \\

            BINALIWALA &
            \includegraphics[width=.3\columnwidth]{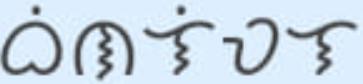} & 
            \includegraphics[width=.3\columnwidth]{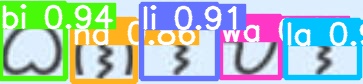} \\

            MAGTAGUMPAY &
            \includegraphics[width=.3\columnwidth]{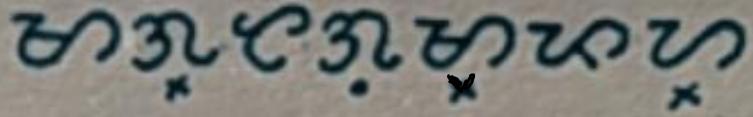} & 
            \includegraphics[width=.3\columnwidth]{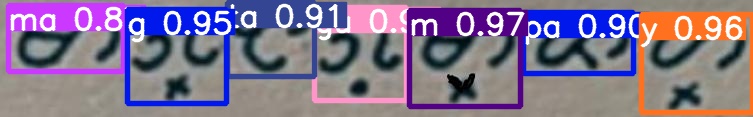} \\
            
            \end{tabular}
        \caption{OAssuming an input image of \emph{Baybayin} (second column), our proposed model performs end-to-end character instance detection, i.e. localization and classification, of detected character (third column).}
        \label{tab:image_pre_processing}
    \end{figure}
\end{center}

\section{Introduction}

\emph{Baybayin} is a Philippine script and writing system that lost its usefulness after being replaced by the now prominent Latin alphabet sometime during Spanish colonization. It is an \emph{Abugida} writing system that utilizes fourteen consonants and three vowels wherein each symbol in the alphabet consists of a consonant followed by a vowel \cite{Cabuay2009}. It shares similarities with Thai and Sanskrit scripts and other \emph{Abugida} writing systems. These three falls under the Brahmic family of scripts. 

Recently, the \emph{National Writing System Act} \cite{houseofrepresentatives} was passed in 2018 to promote the use of \emph{Baybayin} in the Philippines to preserve its cultural and historical importance to the country. Although the bill has already been passed, criticism regarding the adoption of \emph{Baybayin} surfaced, especially where there could be difficulty in learning the writing system today, which is already the Latin alphabet-dominated Filipino writing system. This paper proposes a model that uses a Convolutional Neural Network (CNN) for \emph{Baybayin} character instance detection using YOLOv8. The proposed model will classify written \emph{Baybayin} text into its Latin counterpart to pave the way for more accessible learning of the writing system. Unlike other systems, our proposed model performs character instance detection in an end-to-end manner as well as extending \emph{Baybayin} classification to its modernized version, where diacritics are present.

\section{Related Works}

Nogra et al. \cite{nogra2020baybayin}, Pino et al. \cite{pino2021baybayin, pino2021optical}, and Bague et al. \cite{bague2020recognition} have previously worked on machine learning projects that aim to provide a model that classifies \emph{Baybayin} characters. Each uses different classification models, including the use of Convolutional Neural Networks (CNNs) \cite{nogra2020baybayin, bague2020recognition}, Support Vector Machine (SVMs) \cite{pino2021baybayin, pino2021optical}, and Long-Short-Term Memory (LSTM) models \cite{NograBaybayinLSTM2019}. However, most of these systems assume that inputs are per \emph{Baybayin} character rather than whole \emph{Baybayin} words, which would require character instance segmentation.

Most existing models \cite{NograBaybayinLSTM2019, pino2021baybayin, pino2021optical} use more traditional machine learning techniques to perform \emph{Baybayin} character classification. Because of the strict pipeline of feature extraction for traditional models, it is generally harder for them to generalize. Hence, limiting their operation to only controlled or semi-controlled image settings \emph{i.e.} background, font, and image quality. Rizky et al. \cite{rizky2018mbojo} worked on a project that employs the use of shearlet transform feature extraction and SVM for \emph{Mbojo} character recognition. \emph{Mbojo} script is similar to \emph{Baybayin}, wherein each consonant letter inherently has a vowel attached to it, and adding punctuation (diacritics) changes the vowel attached to it. The inaccuracies of the studies are reported to stem from the pre-processing lacking, and an improvable feature extraction method. The incorporation of automated feature extraction using CNNs or other deep-learning techniques may solve this issue. 

Deep learning-based models \cite{bague2020recognition, nogra2020baybayin} on the other hand perform feature extraction automatically as part of their internal pipeline, hence removing external limits for manual feature extraction and its potential "strictness". A study that sought to compare the performance of traditional machine learning models (SVM) and deep learning models (YOLO-based CNN) \cite{Wang2019ComparisonOfAlgorithms} concluded that despite deep learning models taking up slightly more computational power and time, they are able to generalize more and reliably provide accurate predictions despite contamination such as occlusions. However, with the emergence of state-of-the-art detection and segmentation CNNs, required computational complexity and inference time continue to decrease significantly.




Moreover, existing work that did use deep-learning techniques for \emph{Baybayin} classification \cite{nogra2020baybayin, bague2020recognition} used datasets where images are either synthesized (computerized font, controlled background) or handwriting in a relatively controlled setting (clear contrast between foreground and background, written well, proper spacing). This limits their performance as a result of a lack of model resilience and generalization. There is limited existing work that can segment and classify \emph{Baybayin} characters outside a controlled environment \emph{i.e.} poor lighting conditions, present occlusions, poor image quality, and much more. 

To our knowledge, this is the first study that features end-to-end character segmentation and classification from modernized \emph{Baybayin} words using a state-of-the-art deep-learning system. This way, a lightweight and real-time inference system can be achieved, and thus is suitable for integration into production applications that will aid in learning the writing system that House Bill 1022 aims to promote and preserve.

\section{Methodology}


This study comprises various experiments to achieve the optimal model for the lightweight deployment of an OCR system. The You-Only-Look-Once V8 (YOLOv8) model was used as the primary machine learning component of the pipeline, with interventions done during the data preprocessing and augmentation stages and experimentation with different parameter counts for the model.


\begin{figure}[t!]
    \centering
    \includegraphics[width=\columnwidth]{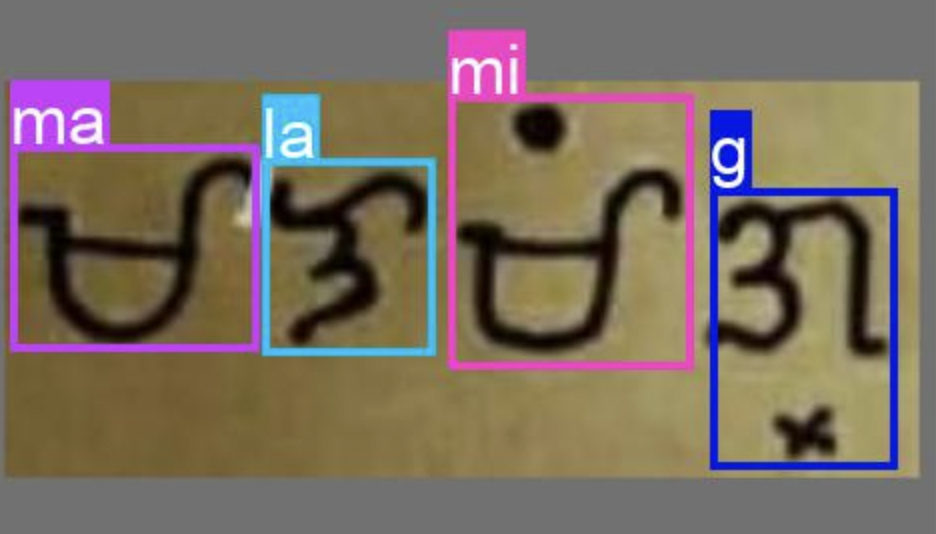}
    \caption{Sample annotated image, "malamig". Annotations on the images are on the character level, where each character has its Latin counterpart as its classification label. Diacritics, like the ones seen in characters "\emph{mi}" and "\emph{g}" are labeled together with their parent character.}
    \label{fig:sample_annotation}
\end{figure}

\begin{center}
    \setlength{\tabcolsep}{1pt}
    \begin{figure*}[t!]
        \begin{tabular}{ccccc}
        Input & Grayscale & Sharpen & Denoise & Binarize \\
        \includegraphics[width=.19\linewidth]{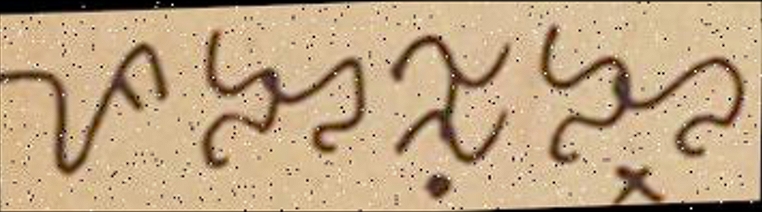} & \includegraphics[width=.19\linewidth]{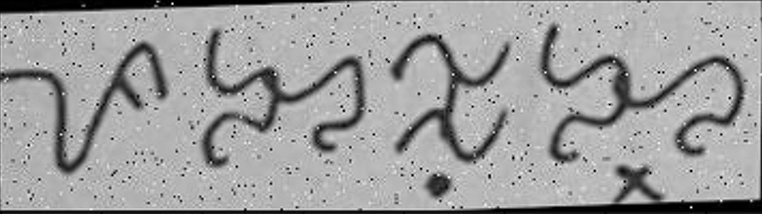} & \includegraphics[width=.19\linewidth]{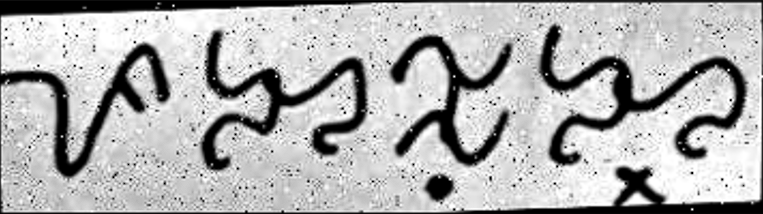} & \includegraphics[width=.19\linewidth]{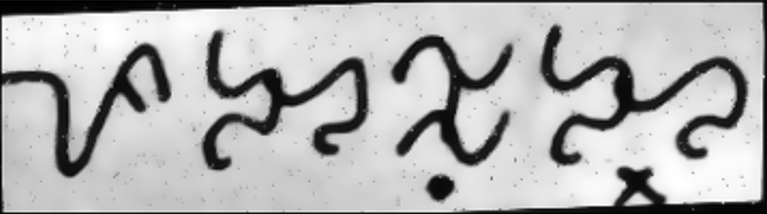} & \includegraphics[width=.19\linewidth]{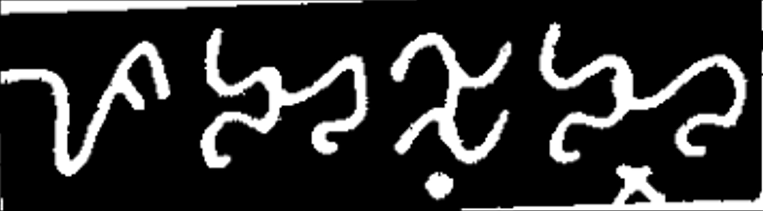} \\
        \includegraphics[width=.19\linewidth]{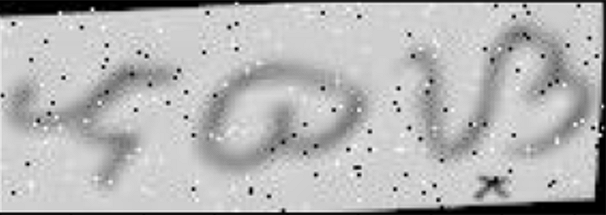} & \includegraphics[width=.19\linewidth]{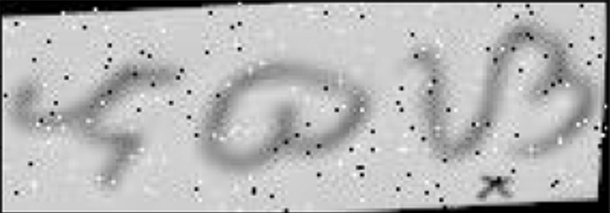} & \includegraphics[width=.19\linewidth]{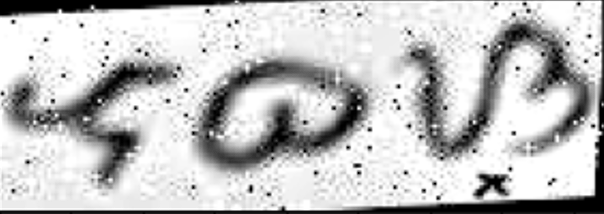} & \includegraphics[width=.19\linewidth]{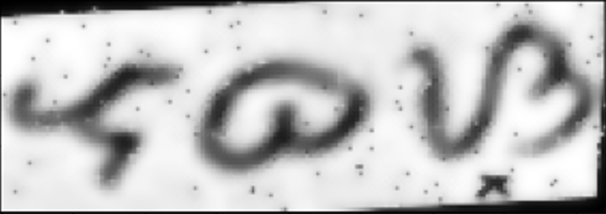} & \includegraphics[width=.19\linewidth]{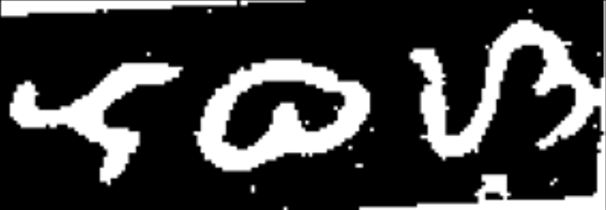} \\
        \includegraphics[width=.19\linewidth]{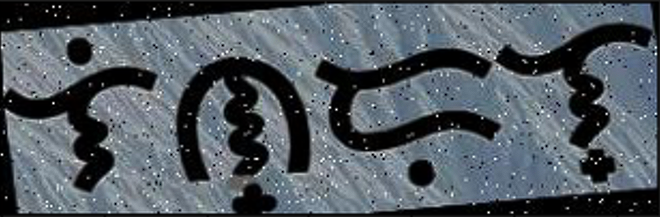} & \includegraphics[width=.19\linewidth]{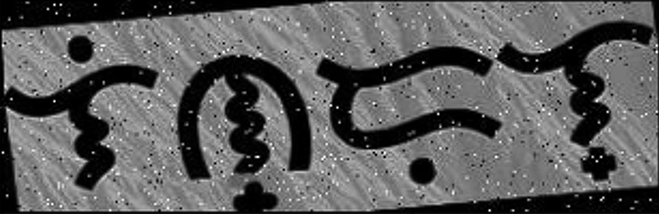} & \includegraphics[width=.19\linewidth]{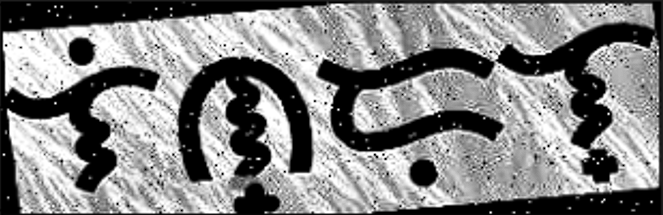} & \includegraphics[width=.19\linewidth]{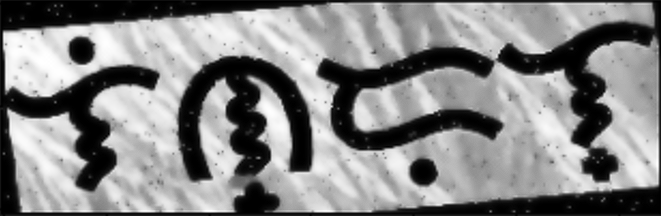} & \includegraphics[width=.19\linewidth]{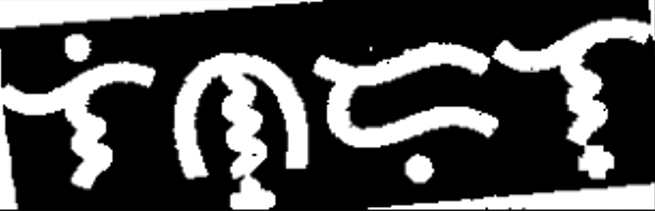} \\
        \includegraphics[width=.19\linewidth]{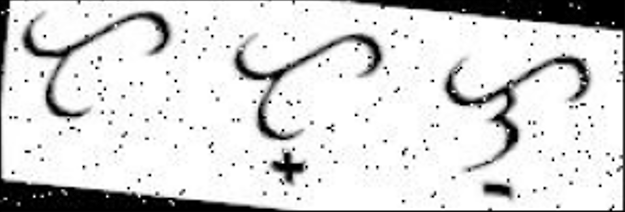} & \includegraphics[width=.19\linewidth]{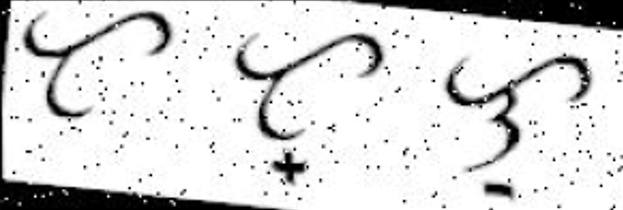} & \includegraphics[width=.19\linewidth]{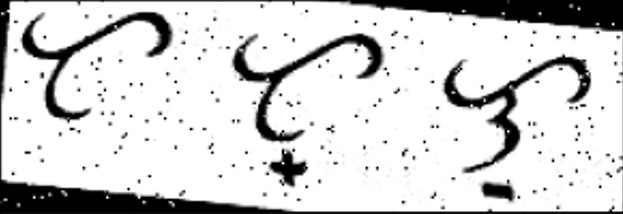} & \includegraphics[width=.19\linewidth]{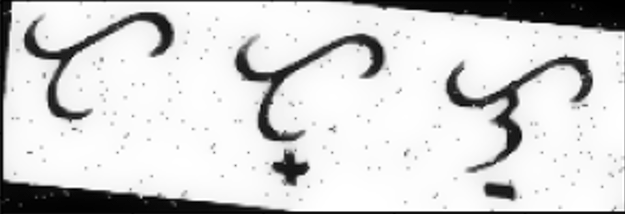} & \includegraphics[width=.19\linewidth]{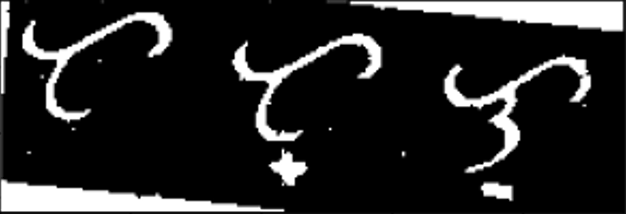} \\
        \includegraphics[width=.19\linewidth]{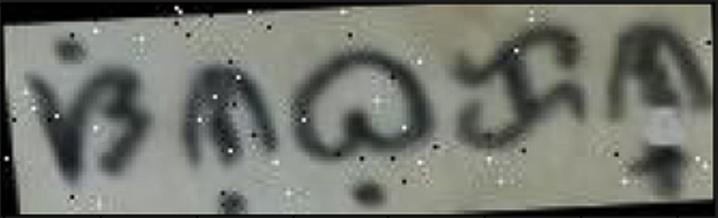} & \includegraphics[width=.19\linewidth]{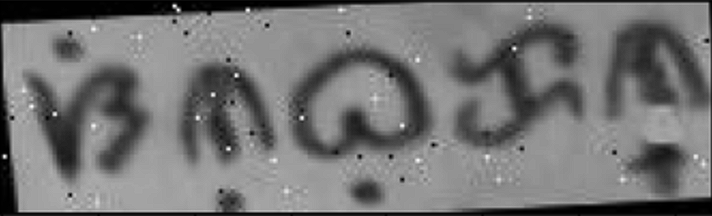} & \includegraphics[width=.19\linewidth]{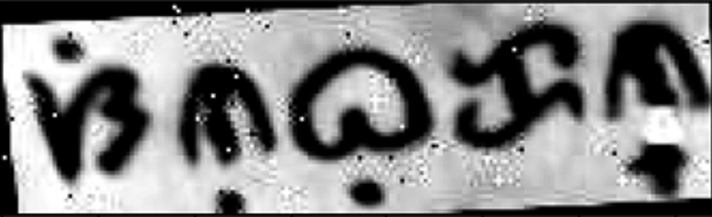} & \includegraphics[width=.19\linewidth]{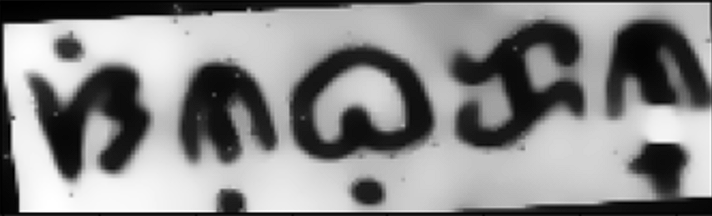} & \includegraphics[width=.19\linewidth]{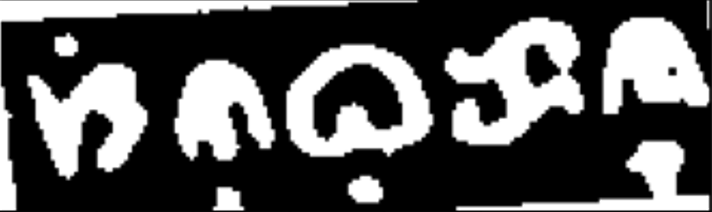}
        \end{tabular}
    \caption{Pre-processing pipeline for experiments that need image transformation. Images are processed from left to right, each row corresponds to a sample image. The binarized images are used as input images for experiments that involve pre-processing.}
    \label{tab:image_pre_processing}
\end{figure*}
\end{center}

\subsection{Dataset}

A new dataset was built in this study, comprising a combination of existing \emph{Baybayin} image datasets already found online. Images were primarily taken from Pino et al. \cite{pino2021baybayin} and Nogra et al. \cite{nogra2020baybayin}, both of whom have published their own \emph{Baybayin} datasets on Kaggle. Both of these studies' datasets were used as they contained handwritten images and were not synthesized. This is important as it makes the final model able to discern handwritten characters from varying backgrounds. The final unprocessed and non-augmented dataset yielded $1001$ images of \emph{Baybayin} words. See Figure \ref{fig:sample_annotation} for an annotated example of the dataset contents.

Due to the use of an end-to-end deep learning model, emphasis is given to data quality and quantity. With $n=1001$, it is barely enough samples to train an end-to-end system. Hence, we perform data augmentation to increase the number of images in the dataset and to inject diversity to increase the model's overall resilience against common environmental phenomena such as skewing, rotation, occlusions, shear, saturation, exposure, and image noise, as these were the augmentation operations applied.

For effective detection and segmentation, proper labels must be in place in all images to establish ground truth. Substantial consideration was placed into different approaches for labeling, especially with the presence of diacritics in the \emph{Baybayin} script. Diacritics are additional markings that augment the original \emph{Baybayin} characters that were added during the modernization of the script. An example of which can be seen in Figure \ref{fig:sample_annotation}, in the rightmost character (Latin "g"). Without the diacritic below the main character, it would be classified as "ga" instead of "g." The dot above the "mi" character is also an example of a diacritic.

For this study, diacritics are considered unified with the main character and are labeled as such. This is possible with \emph{Baybayin} as it is an alphabet-based language, unlike Chinese or Japanese Kanji, where new characters may be created based on the combinations of radical characters. See figure \ref{fig:sample_annotation} for annotation samples.

\subsection{Pre-Processing}
\label{sec:preprocessing}

For certain experiments, data is first put through a pre-processing pipeline to remove or emphasize certain image features before training occurs. The Scikit-image and OpenCV-Python libraries were employed to facilitate these pre-processing techniques.

The RGB images are first converted to grayscale to avoid color non-uniformity issues. They speed up model performance by giving the model only a single channel of color information instead of three. The images are then sharpened to define the edges further to aid in edge detection for better segmentation and character detection. For more successful character recognition, the distinction between the object in the foreground and the background is key. To achieve this, binarization was applied to the sharpened images. The thresholding technique used was Otsu's Binarization. The technique computes a threshold value by finding the least weighted variance between the foreground and the background. After binarization, noise removal is applied to the images using a total variation process. Lastly, normalization is done to transform the images to similar dimensions for dataset uniformity.

\subsection{Training and Evaluation}

A YOLOv8 architecture was used to perform object detection and segmentation on the \emph{Baybayin} dataset. We performed experiments both training the architecture from scratch, as well as using YOLOv8 pre-trained on the COCO dataset. Also, experiments were done on the YOLOv8 Medium, Small, and Nano configurations, where medium, small, and nano reference the number of parameters relative to each other. The medium configuration was used as a baseline to assess the model's performance with the average number of parameters, whereas the small and nano configurations were used to determine the model's performance for lightweight applications.

For all architecture configurations, ReLU (detector) and Softmax (classifier) were considered as the activation functions with Stochastic Gradient Descent (SGD) as the optimizer, minimizing the Binary Cross-Entropy Loss with Logits. Each iteration used 32 images with input images resized for uniformity to $640 \times 640$ px. With an initial learning rate of $0.01$, and SGD momentum of $0.937$, the training spanned for a maximum of 600 epochs, but a patience threshold of 100 epochs was enforced such that if no improvements are observed within the said amounts of epochs, training stops early. Additionally, weighted-class training is also implemented to alleviate class imbalances in the dataset.


There are 2,400 images for training, 100 images for validation, and 100 images for testing. Output metrics from testing such as the accuracy, mAP, F1 score, and loss curves were used to evaluate the trained model, combined with a visual inspection of resulting model predictions. All model training experiments were done using a computer system equipped with an Nvidia RTX 3090 GPU with 24GB of video memory running on the \emph{pytorch/pytorch} Docker image.

\subsection{Experiments}

In addition to the creation of an end-to-end system for \emph{Baybayin} segmentation and classification and studying its performance across multiple parameter counts, there is also merit in experimenting with whether or not applying pre-processing techniques (see section \ref{sec:preprocessing}) impacts the performance and operation of the system. As such, for each variant of the YOLOv8 architecture (nano, small, and medium), a model was trained on a version of the dataset that has not been pre-processed, and another model was trained on a version of the dataset that has been pre-processed. All the said experiments were done using the full augmented dataset. As such, a version of the augmented dataset with the pre-processing applied was created. Subsequently, a version of the augmented dataset was retained without the pre-processing steps applied.

Similar to training, all experimental inferences were done on an NVidia RTX 3090 GPU with 24GB VRAM. In addition, to study performance on systems without access to GPU, experimental inferences were also done using CPU-only, specifically using an AMD Ryzen Threadripper with 16GB of memory. Performance metrics such as average inference or prediction time, average frame rate for video stream inference, mAP, accuracy, and F1-Score were used for model comparison.

\section{Results \& Discussions}

\subsection{Model Training and Evaluation}


\begin{figure}[t!]
    \centering
    \includegraphics[width=\columnwidth]{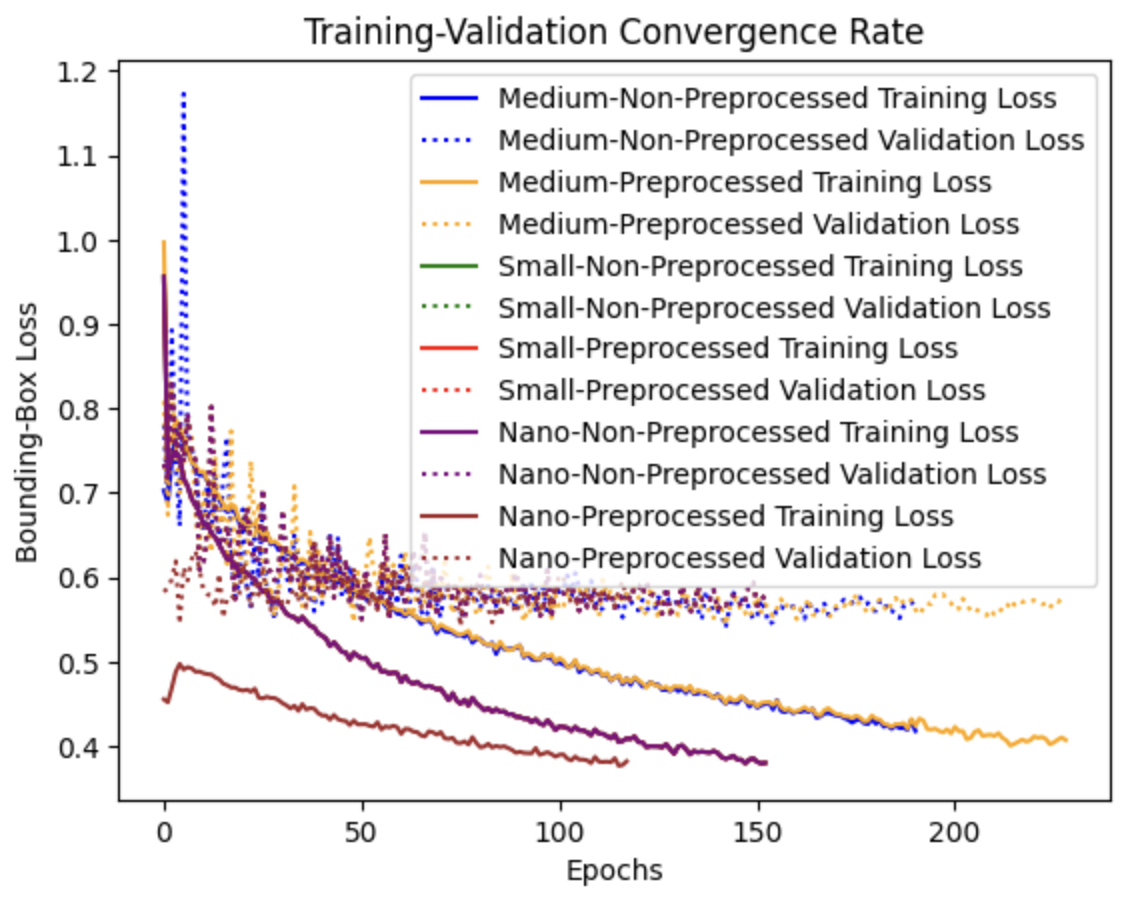}
    \caption{Training and Validation Loss. The disparity in convergence rates is highlighted across different parameter counts. The Medium models converge much later than the rest. Both the Small models and the Nano-non-preprocessed model share identical loss profiles. The Nano-preprocessed model started out with the least loss, leading also for it to converge the earliest. Validation loss never went below 0.55.}
    \label{fig:training_convergence}
\end{figure}

As seen in Figure \ref{fig:training_convergence}, all models, regardless of parameter count, ceased to improve around the 75th epoch or earlier. From that point onwards, the training loss kept decreasing while the validation loss kept fluctuating but keeping the same plateau trend. This could signify a bottleneck in the dataset itself since despite the continued decrease in training loss, the validation loss fluctuates around a fixed range, not creating an increasing trend. This means that no overfitting is taking place, but no features are being learned either.


\begin{table*}[!ht]
    \centering
    \caption{Performance and Validation metrics. Arranged by parameter count, nano models have the least parameter counts to the medium models, which have the most parameter counts for this series of experiments. The mAP@50 and mAP@50-95 are logged, but the use of the F1-Score for model evaluation is preferred due to class imbalance in the dataset.}    
    \begin{tabular}[width=\textwidth]{llllllll}
    \hline
        \textbf{Model} & \textbf{GPU Inference Time} & \textbf{CPU Inference Time} & \textbf{mAP@50} & \textbf{mAP@50-95} & \textbf{Precision} & \textbf{Recall} & \textbf{F1-Score} \\ \hline
        N-Non-Preprocessed & 2.1 ms/img & 83.2 ms/img & 0.892 & 0.775 & 0.879 & 0.802 & 0.8387 \\ 
        N-Preprocessed & 2.1 ms/img & 85.3 ms/img & 0.897 & 0.778 & 0.859 & 0.809 & 0.8333 \\ 
        S-Non-Preprocessed & 2.1 ms/img & 84.3 ms/img & 0.892 & 0.775 & 0.879 & 0.802 & 0.8387 \\ 
        S-Preprocessed & 2.1 ms/img & 82.4 ms/img & 0.892 & 0.775 & 0.879 & 0.802 & 0.8387 \\ 
        M-Non-Preprocessed & 3.3 ms/img & 189.2 ms/img & 0.933 & 0.805 & 0.858 & 0.839 & 0.8484 \\ 
        M-Preprocessed & 3.3 ms/img & 196.3 ms/img & 0.922 & 0.802 & 0.854 & 0.857 & 0.8555 \\ \hline
    \end{tabular}
    \label{tab:validation_metrics}
\end{table*}

All models are performing generally well. With the highest attained F1-Score of $0.856$ by the pre-processed medium model. Paired with a mAP@50-95 of $0.857$, it proves to be a robust model for \emph{Baybayin} segmentation.

\begin{figure}[t!]
    \centering
    \includegraphics[width=\columnwidth]{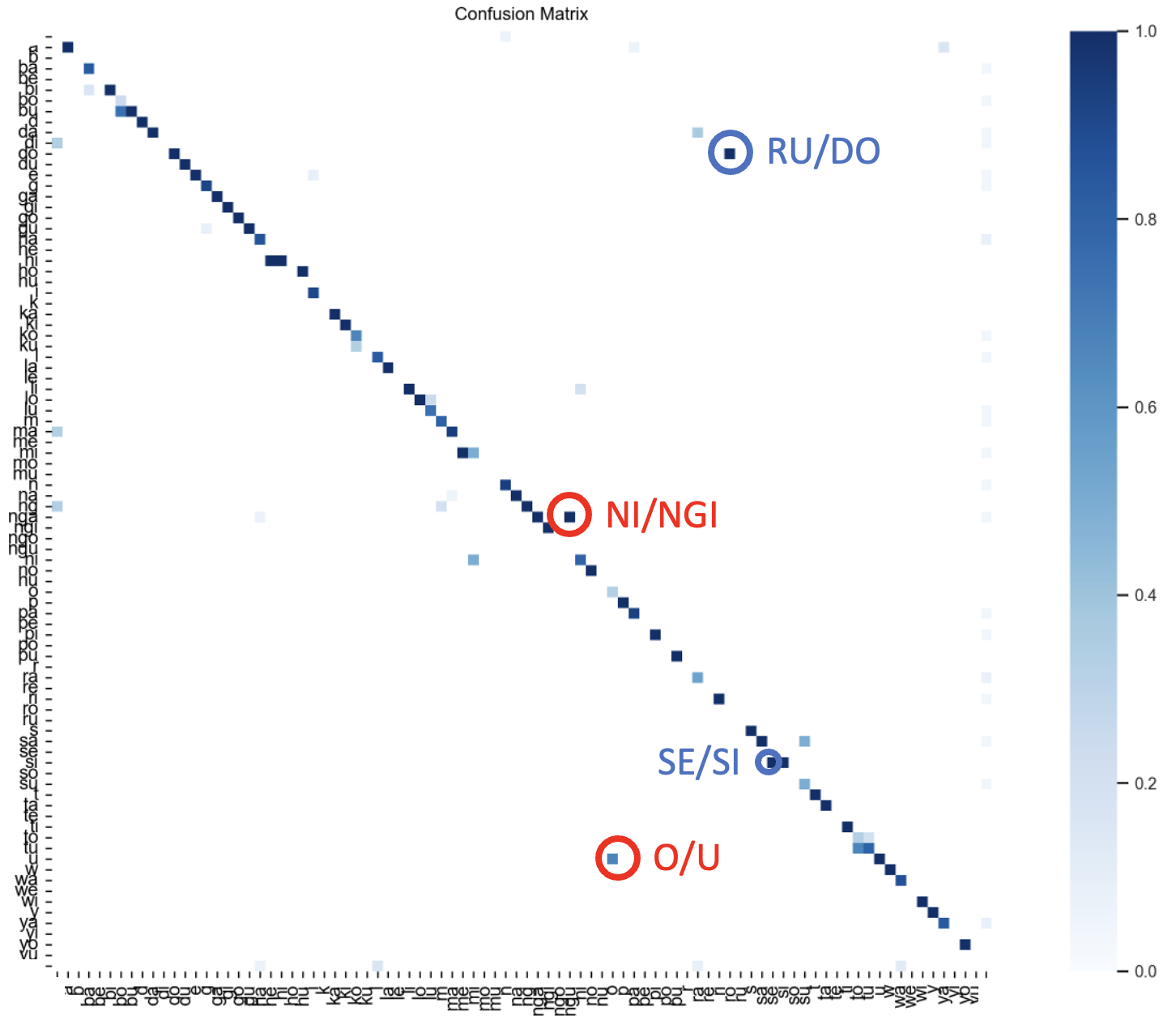}
    \caption{Classifier confusion matrix. Most hits in the matrix belong to the main diagonal, which indicates the correct classification between the ground truth Latin counterpart and the predicted Latin counterpart of the \emph{Baybayin} character. Annotated in blue are examples of misclassifications due to translation ambiguity while annotations in red are examples of misclassifications due to diacritic misclassification.}
    \label{fig:classification_confusion_matrix}
\end{figure}

The confusion matrix in Figure \ref{fig:classification_confusion_matrix} shows the distribution of all predictions during the model validation stage. It can be seen that there are a few outliers that stray from the main diagonal of the matrix, indicating a misclassification. The misclassification generally comes from \emph{Baybayin} characters that may have more than one counterpart in the Latin script, causing translation ambiguity. Some examples of which are \emph{RA/DA}, \emph{DO/RO}, \emph{O/I}, and \emph{E/I}. Another source of confusion for the system is the way that diacritics are positioned and shaped. Depending on the diacritic shape and location, it may change the appended vowel to the character.

\begin{figure*}[t]
    \centering
    \begin{tabular}{c}
         \includegraphics[width=\textwidth]{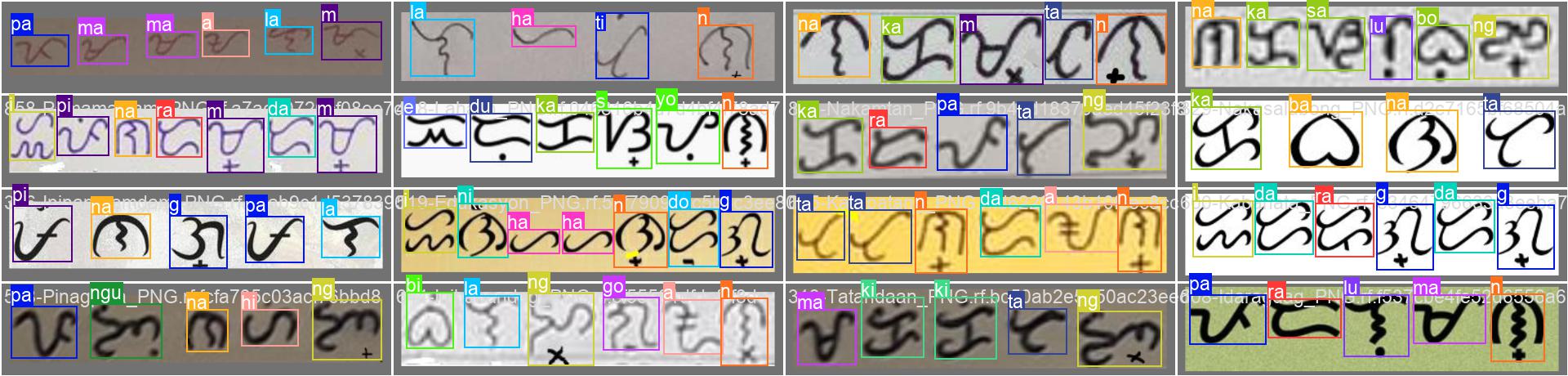}  \\
         \includegraphics[width=\textwidth]{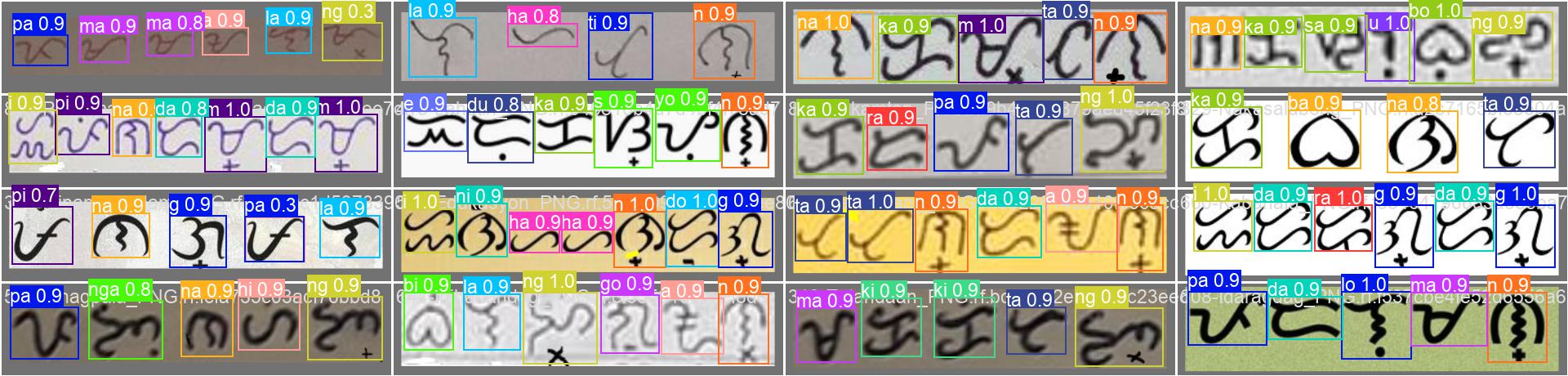}
    \end{tabular}
    \caption{Validation mosaic. These are sample predictions made on the validation split of the dataset. The labels are indicated in the upper region of this figure-table, while the predictions with their corresponding confidence scores are shown in the lower region of this figure-table.}
    \label{tab:sample_validation_predictions}
\end{figure*}

\begin{figure*}[t!]
    \centering
    \includegraphics[width=\textwidth]{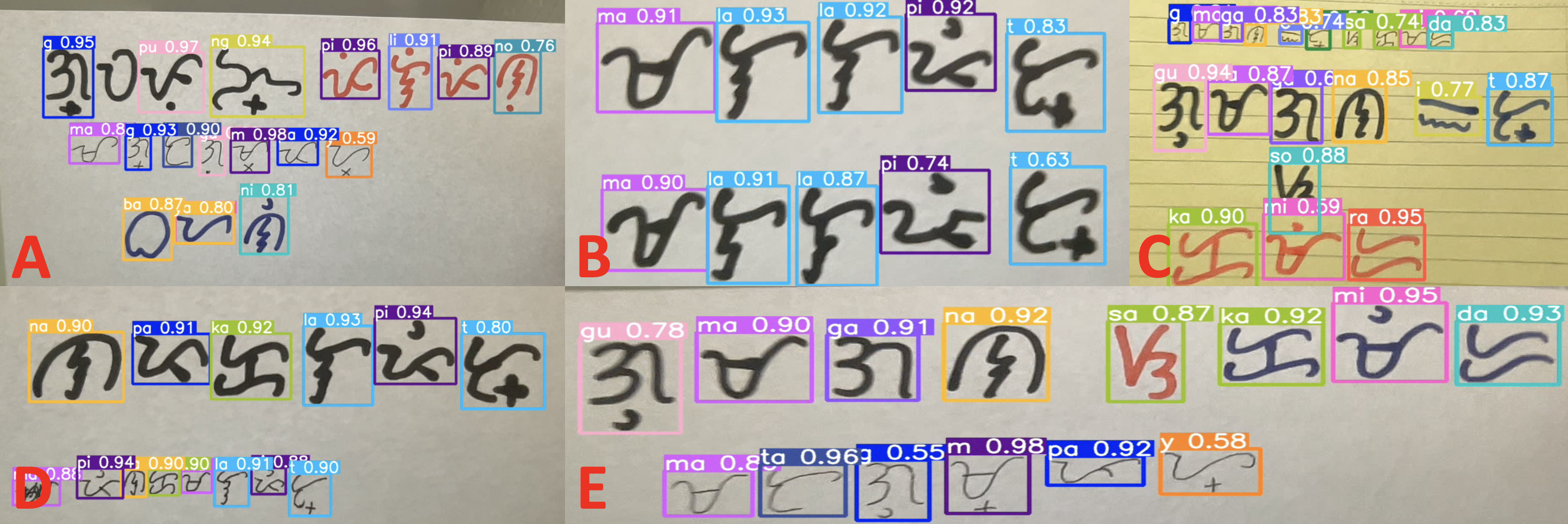}
    \caption{Predictions using camera hardware. A) Figure shows model resilience against different stroke colors. B) Figure shows ability to segment characters that are slightly close to each other in terms of distance. C) Figure shows resilience against colored backgrounds, different stroke colors, and background artifacts like the gridlines. D) Figure demonstrates the ability to segment tightly-compacted characters. E) Figure shows classification performance.}
    \label{fig:camera_predictions}
\end{figure*}

\subsection{Difference in Parameter Counts}


As observed in Table \ref{tab:validation_metrics}, the medium models outperformed all of the small and nano models. This is due to its higher parameter count, which allows for more features to be learned. However, it is interesting to point out that the small and nano models performed nearly identically to each other, which could raise some assumptions about the architecture of these model configurations. It is possible that parameters were reduced for some layers in the convolutional model in such a way that these parameters did not play much of a significant role when placed in the context of this study, such that the small and nano models did not need those pruned parameters. It is also possible that the parameter differences between nano and small were not enough to extract the features that could have been extracted if the medium model were to be used, leading to the said parameters to zero out in the end-to-end pipeline. This may happen if the layers pruned from small to nano architectures were not layers that perform feature-extraction itself, such as filters and not kernels.

Additionally, applying pre-processing on the dataset prior to training did not yield any significant difference in validation metrics and inference performance. This highlights that the features learned by the system are more on shapes and lines rather than colors.

Given that performance gaps between the different architecture configurations are not significantly far apart, the smaller configurations are robust enough to be suitable for use on devices with lower computational power.




\subsection{Inference Performance and Applications}


Figure \ref{fig:camera_predictions} shows examples of predictions from a live web camera stream. When running on camera hardware, the model proves to be accurate and resilient enough to provide reliable segmentation and classification. Figure \ref{fig:camera_predictions}a shows that different stroke colors are acceptable to the model as long as its contrast is reasonably different from the background. Further, Figure \ref{fig:camera_predictions}c shows the same effect, with added resilience against a different colored background and with a grid line. Figures \ref{fig:camera_predictions}b and \ref{fig:camera_predictions}d demonstrates that the model can reliably segment the spatially-compressed characters. Lastly, Figure \ref{fig:camera_predictions}e highlights classification reliability, which shows that the system presents sufficient accuracy and reliability for a production application.

    
With a GPU inference time of 3.3 ms per frame (see Table \ref{tab:validation_metrics}), there is a capacity to process 303 frames per second (FPS) for a GPU similar to the Nvidia RTX 3090. As video stream standards are around the range of 30 FPS to 120 FPS, a yield of 303 FPS provides a large performance margin for weaker GPUs to fill in. Such GPU applications of this work would focus on translating large amounts of characters in a relatively small amount of time.

    
Unlike the GPU, the CPU inference time takes significantly longer at 196.3 ms per frame. This yields a capacity to process only 5 FPS. However, downscaling the model to the small or the nano architecture configurations will allow the model to perform at around 84 ms per frame, which yields a capacity to process approximately 12 FPS, which is more than 50\% the CPU performance using the medium architecture configuration. 12 FPS is acceptable for use in everyday production deployments such as smartphone live video stream OCR.

\subsection{Issues and Failure Cases}

The performance of the segmentation stage and classifier stage in tandem can be seen in Table \ref{tab:sample_validation_predictions}. Shown here are mostly rigidly handwritten images of the \emph{Baybayin} script in the validation split of the dataset. As discussed earlier, there are some misclassifications regarding vowels. For example, the first-column fourth-row image has \emph{NGU} labeled in green. This character was predicted to be \emph{NGA} instead. Similarly, the first-column second-row image has \emph{RA} labeled in red. This character was predicted to be \emph{DA} instead. Issues like these cannot directly be resolved by providing intervention to the classifier or segmentation stage. This is because in Filipino grammar, the use of these different pronunciations of the same character is decided by the context of the sentence or the context of the word. In this light, a contextualization stage could be added, similar to what was done in \cite{pino2021baybayin}, where they used a Filipino word list to pinpoint the closest possible Filipino word in Latin text that would best translate the \emph{Baybayin} word script.

\begin{figure}[t!]
    \centering
    \includegraphics[width=\columnwidth]{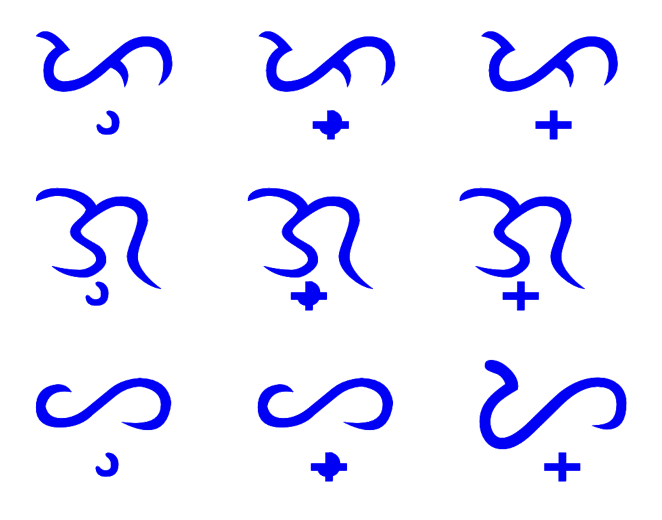}
    \caption{Diacritic similarities. The first row of characters are PU/PO/P, second row are GU/GO/G, third row are YU/YO/Y.}
    \label{fig:diacritics_similarities}
\end{figure}

Moreover, there are still some instances where diacritics are misclassified. Diacritics often look similar, hence can be misclassified. Figure \ref{fig:diacritics_similarities} shows how similar diacritics can be with each other. Paying close attention to the two rightmost characters in the figure, both diacritics bear the same general shape, with the only difference in stroke weight and an additional blob within the cross. Looking at the first character in each row, we can also conclude that the center character's diacritic is a combination of the other two diacritics. This was often misclassified by the system. This is also evident in Figure \ref{fig:classification_confusion_matrix}.

\begin{figure}[t!]
    \centering
    \includegraphics[width=\columnwidth]{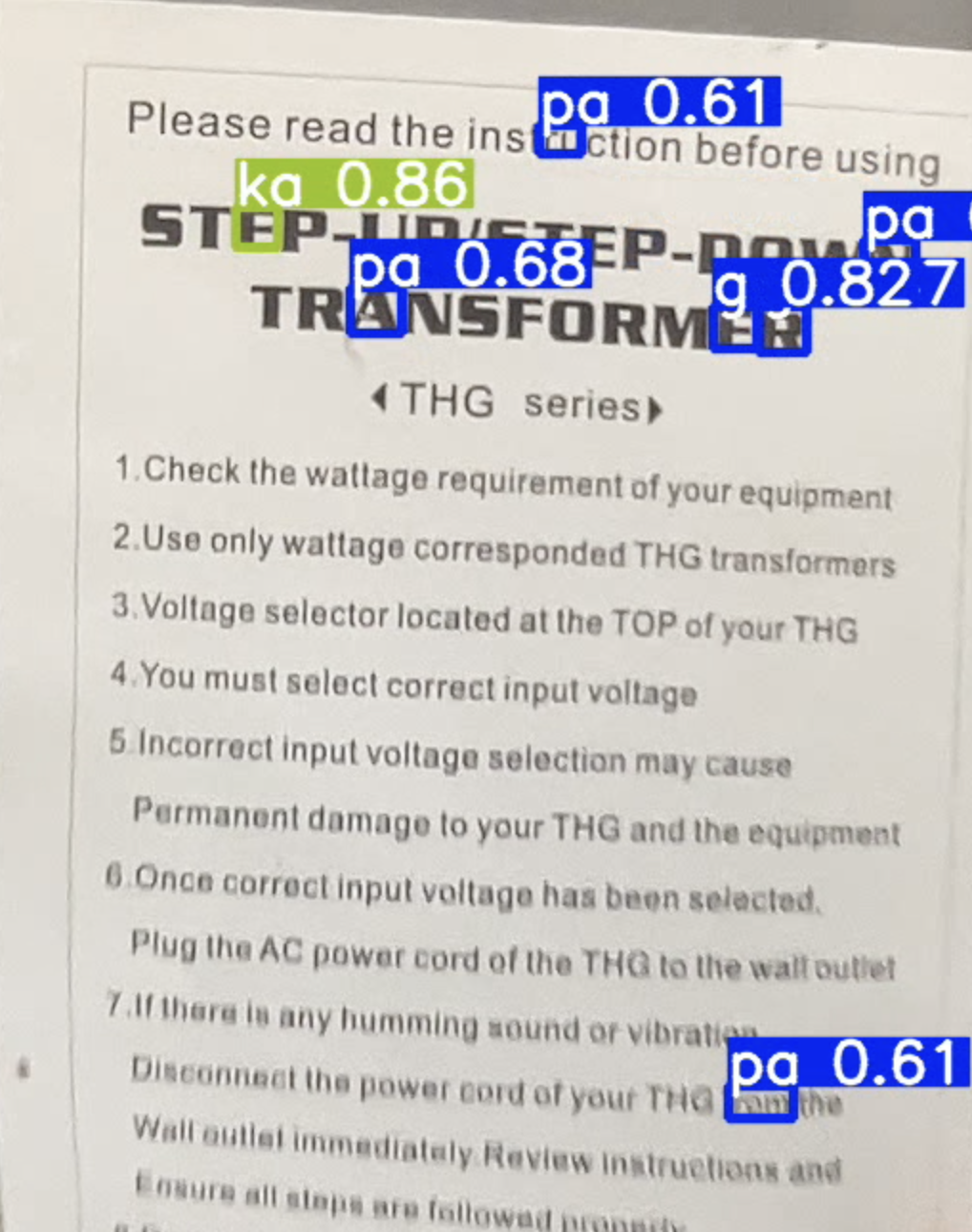}
    \caption{Misclassification of Latin Text. Some Latin text were segmented and misclassified as Baybayin with a confidence threshold of 0.6.}
    \label{fig:latin_misclassifications}
\end{figure}

Additionally, the handling of non-\emph{Baybayin} scripts in input images was not taken into account. Because of this, though \emph{Baybayin} detections, segmentations, and detections are correct, there may be some Latin text that could be misclassified as \emph{Baybayin} text. A common example of this would be the Arabic number \emph{6} and the Latin alphabet character \emph{y}. They are often misclassified for \emph{SA} and \emph{YA} respectively. Examples of such misclassification are found in Figure \ref{fig:latin_misclassifications}. To address this, future iterations of this work should include non-\emph{Baybayin} text in training datasets to avoid misclassifications.

Further, there is the distinct issue of the dataset class imbalance. The effects of class imbalance were mostly mitigated by the use of weighted class training, but this still increases the likelihood of misclassifications to happen.

Lastly, it was evident in Figure \ref{fig:training_convergence} that validation loss remained constant regardless of the training loss. This is an indicator that work should be done on the dataset by increasing the number of images and diversity of images for better generalization, especially in the image setting, quality, and background.

\section{Conclusions and Recommendations}

In summary, we propose a system performing character segmentation on \emph{Baybayin} words. Our system yielded good performance with the highest mAP50 of $0.933$, mAP50-95 of $0.805$, and an F1-Score of $0.848$. To our knowledge, this is the first end-to-end system that performs character segmentation and classification on modern \emph{Baybayin} words with diacritics. Using this approach enabled for a relatively lightweight end-to-end model, with $11,159,253$ parameters or 22 MB in file size. This allows for the deployment of lightweight production applications to promote \emph{Baybayin} by aiding those who cannot easily read the script.

For further iterations of this study, it is recommended to create a much larger, more diverse dataset for better model generalization. Preferably, diverse fonts or writing styles of \emph{Baybayin} should also be included, along with diverse text backgrounds. The addition of non-\emph{Baybayin} classification is also  recommended to avoid the misclassification of non-\emph{Baybayin} objects. Further, context-sensitive post-processing of output predictions should also be done to remove potential ambiguous results that can only be distinguished by context.

\bibliographystyle{ACM-Reference-Format}
\bibliography{sample-base}










\end{document}